\pdfoutput=1

\documentclass[11pt]{article}

\usepackage[final]{acl}

\usepackage{times}
\usepackage{latexsym}

\usepackage[T1]{fontenc}

\usepackage[utf8]{inputenc}

\usepackage{microtype}

\usepackage{inconsolata}

\usepackage{graphicx}

\usepackage{subfigure}
\usepackage{color}
\usepackage{tabularray}
\usepackage{rotating}

\newcommand{\comment}[1]{}

\title{Routing in Sparsely-gated Language Models responds to Context}

\author{Stefan Arnold \and Marian Fietta \and Dilara Yesilbas \\ 
Friedrich-Alexander-Universität Erlangen-Nürnberg \\ Lange Gasse 20, 90403 Nürnberg, Germany \\ 
\texttt{(stefan.st.arnold, marian.fietta, dilara.yesilbas)@fau.de}}

\begin{document}
\maketitle

\begin{abstract}
Language Models (LMs) recently incorporate mixture-of-experts layers consisting of a router and a collection of experts to scale up their parameter count given a fixed computational budget. Building on previous efforts indicating that token-expert assignments are predominantly influenced by token identities and positions, we trace routing decisions of similarity-annotated text pairs to evaluate the context sensitivity of learned token-expert assignments. We observe that routing in encoder layers mainly depends on (semantic) associations, but contextual cues provide an additional layer of refinement. Conversely, routing in decoder layers is more variable and markedly less sensitive to context. 

\end{abstract}

\section{Introduction}

\textit{Language Models} (LMs) have demonstrated exceptional capabilities in capturing linguistic nuances \citep{devlin2019bert} and generating coherent text \citep{radford2019language, brown2020language}. However, the dense nature of their architectures, where each token is processed by the total number of parameters, inherently limits their scalability, which is considered the predominant driver for their advanced expressiveness \citep{kaplan2020scaling}.

Sparsely-gated \textit{Mixture-of-Experts} (MoE) models as developed by \citet{shazeer2017outrageously} and more recently integrated into the transformer architecture \citep{vaswani2017attention} by \citet{lepikhin2020scaling} and \citet{fedus2022switch}, emerged as a promising technique to scale up the parameter count of densely-connected language models \citep{brown2020language}. Beyond language models, this design paradigm was successfully applied to vision models \citep{riquelme2021scaling} and vision-language models \citep{shen2023scaling, lin2024moe}, showcasing its versatility and effectiveness across various tasks.

Unlike applying the same parameters to every token as in dense transformers, the guiding design principle of sparse transformers is to selectively activate a subset of parameters for each token \citep{bengio2013estimating}. Specifically, mixture-of-experts layers operate by incorporating routers and making them learn to dynamically direct tokens to specific parameters, referred to as \textit{experts} \citep{jacobs1991adaptive}. This sparsity routing addresses the scaling issues of dense transformers while maintaining a constant number of computational operations.

Since routing is central to the mixture-of-experts paradigm, most ongoing research is dedicated to identifying and relieving various challenges associated with unstable gates \citep{nie2021evomoe, dai2022stablemoe} and representation collapse \citep{chi2022representation, liu2022gating, do2023hyperrouter}. Other research examined routing patterns \citep{zoph2022st, jiang2024mixtral, xue2024openmoe} to assess how effectively a sparse transformer can leverage its diverse set of experts. By tracing routing decisions across expert layers, \citet{zoph2022st} discovered that expert assignments are less uniform among encoder layers than decoder layers and that meaningful specialization manifests primarily in syntactic properties rather than high-level semantics. \citet{xue2024openmoe} further corroborated that routing is predominantly based on token identities and positions, regardless of context. This finding was termed \textit{context-independent expert specialization} and justified by two observations: (1) tokens are routed to only a few fixed experts, and (2) consecutive token positions prefer similar experts. 


\paragraph{Contribution.}

Given the presumption of context-independent routing, we systematically investigate the context sensitivity of \textit{learned} token-to-expert assignments by exploiting annotated pairs of text from \texttt{WordSim} \citep{finkelstein2001placing}, \texttt{SimLex} \citep{hill2015simlex}, \texttt{SCWS} \citep{huang2012improving}, and \texttt{WiC} \citep{pilehvar2019wic}. We find evidence that routing is responsive to contextual cues, as words in similar contexts are more consistently assigned to the same experts compared to words from different contexts. However, we also observe notable differences among the model components and configurations: (1) context sensitivity is more pronounced in the encoder than the decoder (in line with \citealp{zoph2022st}), and (2) context sensitivity increases with the total number of experts.

\section{Background}

\textit{Mixture-of-Experts} (MoE) has a long history in machine learning, dating back to the principle of adaptive mixtures of local experts \citep{jacobs1991adaptive}. \citet{shazeer2017outrageously} recently introduced sparsely-gated layers by extending the mixture-of-experts paradigm with techniques for conditional computation \citep{bengio2013estimating}. By taking advantage of conditional computation, mixture-of-experts layers enable to scale up the number of trainable parameters while maintaining computational costs.


Building on transformer models \citep{vaswani2017attention}, sparse mixture-of-experts layers can be interleaved with dense layers \citep{fedus2022switch} or upcycled from dense layers \citep{komatsuzaki2022sparse}. Sparse layers typically consists of a router and a fixed number of experts that are structurally identical to standard feed-forward neural networks. The router is responsible for assigning inputs to experts. Each input is projected from its hidden state to the set of experts by multiplication with the router weights, which are learned jointly with the other network parameters. To produce a gradient for the router, the output of the computation is weighted by the corresponding probability of the assignment, since this probability is differentiable. This experts-as-a-layer approach dynamically activates a fixed subset of experts, ensuring that the number of floating-point operations remain constant, regardless of the total number of experts. 


To receive sufficient gradients for learning the router weights, \citet{shazeer2017outrageously} conjectured that sparse mixture-of-experts layers require top-$2$ routing. As such, most implementations of sparse layers rely on two-way routing \citep{lepikhin2020scaling, du2022glam}. However, this assumption is challenged by stable modifications for top-$1$ \citep{fedus2022switch, yang2021m6} and adaptive top-$k$ routing \citep{li2023adaptive}, which allows variable expert assignment based on token complexity.


To promote a balanced distribution of workload, \citet{lepikhin2020scaling} defined a fixed \textit{expert capacity}, which limits the number of tokens each expert can be assigned. The expert capacity is typically specified in the form of a hyperparameter, which acts as a multiplier factor for the expected number of tokens that would be assigned to each expert under a perfect uniform distribution. If the number of tokens assigned to an expert is not enough to fill its capacity, its set of tokens is padded to fill the remaining slots. If the number of tokens assigned to an expert overflows its capacity, the extra tokens are dropped. \citet{gale2023megablocks} addressed the token dropout issue by reformulating the computation in terms of block-sparse operations that efficiently handle the dynamism present in sparse layers.

Since routing determines the token-expert assignments and thus dictates how effectively a model can leverage its set of experts, it is of central importance for the mixture-of-experts paradigm. There are two common classes of assignment algorithms for sparse layers: \textit{token choice} in which tokens are dispatched to top-ranked experts and \textit{expert choice} in which experts select the top-ranked tokens.

\paragraph{Token Choice.} The most common routing strategy is \textit{token choice} \citep{shazeer2017outrageously, lepikhin2020scaling, fedus2022switch}, in which routing decisions are made by greedily selecting the top-scoring experts for each token after projecting their hidden states to the number of experts.

However, the greedy nature of this routing strategy suffers from notorious load imbalance issues that may cause the routers to collapse because experts that are assigned zero tokens no longer receive gradient updates \citep{zhou2022mixture}. To encourage routers to make balanced token-expert assignments, additional adjustments such as \textit{noisy gating} \citep{shazeer2017outrageously} and imposing an auxiliary \textit{load balancing loss} \citep{fedus2022switch} are required. \citet{puigcerver2024sparse} developed a \textit{soft routing} strategy with full differentiability that fills the capacity of experts using a weighted average of tokens. This provides a balanced and dropless mechanism for token-expert assignment.

Compared to the learning-to-route paradigm for routers \citep{shazeer2017outrageously, fedus2022switch}, an alternative strategy is to reformulate the routing algorithm as a linear assignment problem that maximizes token-expert affinity \citep{lewis2021base, clark2022unified} or to eliminate the necessity for routers: \textit{stochastic routing} \citep{zuo2021taming} leverages a consistency regularized loss for stochastic assignment, whereas \textit{deterministic hashing} \citep{roller2021hash} employs a parameter-free assignment algorithm that routes tokens by hashing.

\paragraph{Expert Choice.} Rather than directing tokens to top-scoring experts, \textit{expert choice} as proposed by \citet{zhou2022mixture} has experts independently selecting top-scoring tokens, which guarantees perfect load balancing and allows for flexible allocation.

\section{Methodology}

To illuminate the dynamics of routing with respect to context, we need to detail a sparsely-gated language model and the measurement to assess the degree of sensitivity within the sparse layers.

We employ the \texttt{Switch} \citep{fedus2022switch} transformer model, a sparsely-gated variant of the \texttt{T5} \citep{raffel2020exploring} sequence-to-sequence model, trained on a span corruption objective. This objective involves recovering variable-length contiguous segments masked in text, promoting a deeper understanding of contextual information compared to autoregressive models with dense layers \citep{brown2020language, touvron2023llama} and sparse layers \citep{du2022glam, jiang2024mixtral}. The architecture of the \texttt{Switch} transformer consists of an encoder and a decoder, each comprising six sparse layers that alternate between dense and sparse configurations. Each sparse layer contains a variable number of total experts in $\{8, 16, 32, 64, 128\}$, with a single active expert, where its assignment is managed through token choice routing combined with a load balancing loss. The choice of the \texttt{Switch} transformer model is driven by its variable configurations of experts and its simple routing strategy. By tracing token-expert assignments in the sparsely-gated layers \footnote{We extract softmaxed router logits of word pairs. Since the \texttt{Switch} transformer model uses a variant of byte-pair tokenization \citep{kudo2018sentencepiece}, we aggregate words by mean pooling over subword components.}, we can examine the sensitivity of the routing to similarity and the surrounding context.




\paragraph{Measurements for Similarity.}


To ablate whether routing is adaptive to similarity, we leverage the \texttt{WordSim} \citep{finkelstein2001placing} and \texttt{SimLex} \citep{hill2015simlex} datasets. These datasets contain word pairs with human judgment on their similarity on a scale of $[0,10]$. While \texttt{WordSim} captures broader relatedness in terms of associations, \texttt{SimLex} strictly annotates semantic similarity. For each word pair, we calculate the (layer-wise) \textit{Jensen-Shannon Similarity} (JSS) between the routing probabilities and correlate it with the corresponding similarity annotation using the \textit{Spearman correlation}.

\paragraph{Measurements for Context.}


To examine the influence of contextualization on routing decisions, we adopt the \texttt{SCWS} \citep{huang2012improving} dataset. Unlike \texttt{WordSim} and \texttt{SimLex}, containing word pairs in isolation, \texttt{SCWS} provides human judgments on the similarity of word pairs associated with a context. The inclusion of contextual cues for each word pair makes \texttt{SCWS} particularly suitable for measuring the extent to which context influences token-expert assignments in sparsely-gated language models. We correlate the similarity of the routing decisions for word pairs in \texttt{SCWS} with and without context against the provided similarity annotations.

Since most pairs of words in \texttt{SCWS} have dissimilar words, we further exploit the \texttt{WiC} \citep{pilehvar2019wic} dataset \footnote{Only $8\%$ of the pairs of word in \texttt{SCWS} are identical and their assigned scores are substantially higher than those with different word pairs, \textit{i.e.,} $6.8$ compared to $3.6$ on a scale from $[0,10]$ \citep{pilehvar2019wic}.}. Framed for binary classification, \texttt{WiC} is composed of a target word for which two contexts are provided that were carefully designed to trigger a specific meaning. The goal is to identify if the occurrences of the word within the contexts correspond to the same intended meaning. By comparing the routing activations separate for words from identical and different contexts, we can examine the context sensitivity of routers and identify words which are routed differently based on its contextual usage. This allows us to disentangle the effects of context from associative relationships and provide a more nuanced understanding of how routing in sparsely-gated language models is influenced by context.

\section{Findings}

To examine how consistently sparsely-gated transformers route words based on context, we calculate the similarity between the distributions of experts for word pairs and correlate them with human judgments. We interpret strong correlation coefficients as context sensitivity. Unless otherwise noted, we average the routing similarity across sparse layers.

\subsection{Correlation with Similarity}

We commence with the adaptability of routing decisions to associations in terms of relatedness and semantic similarity. Table \ref{similarity} presents the correlation coefficients grouped by encoder and decoder. 

For the encoder, the averaged correlation values are $0.3078$ and $0.1883$, respectively. These correlations indicate that the routing in sparsely-gated language models depend more on \textit{common concepts} than by \textit{strict meaning}, as evident by correlations for \texttt{WordSim} being consistently higher than correlations for \texttt{SimLex} across most numbers of experts. We further notice diminishing returns in routing similarities concerning the total number of experts, as evident by growing scores between $8$ and $32$ experts and a significant drop at $64$ and $128$ experts. This implies certain fluctuations \citep{dai2022stablemoe} when a large number of experts is set.

\begin{table}
\centering
\caption{\small 
    Correlation of routing probabilities with annotations of association and semantic similarity. Annotations for association were derived from \texttt{WordSim}, whereas annotations for semantic similarity were derived from \texttt{SimLex}.
}
\label{similarity}
\resizebox{\linewidth}{!}{%
\begin{tblr}{
  row{odd} = {c},
  row{2} = {c},
  row{4} = {c},
  row{6} = {c},
  cell{1}{1} = {r=2}{r},
  cell{1}{2} = {c=2}{},
  cell{1}{4} = {c=2}{},
  cell{3}{1} = {r},
  cell{4}{1} = {r},
  cell{5}{1} = {r},
  cell{6}{1} = {r},
  cell{7}{1} = {r},
  cell{8}{2} = {c},
  cell{8}{3} = {c},
  cell{8}{4} = {c},
  cell{8}{5} = {c},
  hline{1,9} = {-}{0.08em},
  hline{2} = {2,4}{},
  hline{2} = {3,5}{r},
  hline{3} = {-}{0.05em},
  hline{8} = {-}{},
}
\begin{sideways}Experts\end{sideways} & Encoder &  & Decoder & \\
 & Association & Similarity & Association & Similarity\\
8 & \textbf{0.2804} & 0.1679 & \textbf{0.0699} & 0.0510\\
16 & \textbf{0.3339} & 0.2070 & \textbf{0.1266} & 0.1179\\
32 & \textbf{0.4333} & 0.2706 & \textbf{0.1879} & 0.1127\\
64 & \textbf{0.3513} & 0.1485 & \textbf{0.2435} & 0.1788\\
128 & 0.1403 & \textbf{0.1474} & 0.0690 & \textbf{0.1317}\\
Avg. & \textbf{0.3078} & 0.1883 & \textbf{0.1394} & 0.1184
\end{tblr}
}
\end{table}

For the decoder, the average correlation values are $0.1394$ and $0.1184$, respectively. Compared to routing in the encoder, the consistent yet relatively low correlations in the decoder across all configurations imply that the decoder is generally less adapted for similarity. This is particularly evident from the more modest peaks and the lack of a significant drop-off in correlation values, which indicates less pronounced expert specialization. This observation is consistent with the finding of \citet{zoph2022st} that routing is uniformly distributed.

\begin{table}
\centering
\caption{\small 
    Correlation of routing probabilities of word pairs with and without contextual cues to annotations of \texttt{SCWS}.
}
\label{context}
\resizebox{\linewidth}{!}{%
\begin{tblr}{
  row{odd} = {c},
  row{2} = {c},
  row{4} = {c},
  row{6} = {c},
  cell{1}{1} = {r=2}{r},
  cell{1}{2} = {c=2}{},
  cell{1}{4} = {c=2}{},
  cell{3}{1} = {r},
  cell{4}{1} = {r},
  cell{5}{1} = {r},
  cell{6}{1} = {r},
  cell{7}{1} = {r},
  cell{8}{2} = {c},
  cell{8}{3} = {c},
  cell{8}{4} = {c},
  cell{8}{5} = {c},
  hline{1,9} = {-}{0.08em},
  hline{2} = {2,4}{},
  hline{2} = {3,5}{r},
  hline{3} = {-}{0.05em},
  hline{8} = {-}{},
}
\begin{sideways}Experts\end{sideways} & Encoder     &                 & Decoder     &            \\
        & w/o Context & w/ Context        & w/o Context & w/ Context \\
8       & 0.2439     & \textbf{0.3183}         & 0.1497      & \textbf{0.1531}     \\
16      & 0.3493      & \textbf{0.4050}            & 0.1981      & \textbf{0.2118}     \\
32      & 0.3873      & \textbf{0.4634}            & \textbf{0.2997}      & 0.1519     \\
64      & 0.2562      & \textbf{0.3980}            & 0.2827      & \textbf{0.3761}     \\
128     & 0.1500      & \textbf{0.3079}            & 0.1382      & \textbf{0.2560}     \\
Avg.    & 0.2773    & \textbf{0.3785}           & 0.2137        & \textbf{0.2298}   
\end{tblr}
}
\end{table}

\subsection{Correlation with Context}

We continue with the response of routing decisions to context. Table \ref{context} presents correlation coefficients for both encoder and decoder components with and without contextual embedding.

For the encoder, the correlation coefficients without context range from $0.1500$ to $0.3873$, with an average value of $0.2773$. This indicates a modest correlation, confirming that even without contextual cues, the routing decisions are influenced to some extent by similarity. When context is added, the correlation coefficients range from $0.3079$ to $0.4634$, with an average value of $0.3785$. This significant increase in average correlation indicates that contextual cues enhance routing decisions, allowing the language model to capture similarities among words more effectively.

Although the average correlation in the decoder increases only slightly from of $0.2137$ to $0.2298$ with context, this apparent insensitivity to context is caused by notable variations in the expert configuration. With few experts, such as $8$ and $16$, routing decisions are hardly influenced by contextual cues. However, a larger number of experts, specifically $64$ and $128$, demonstrates that context can substantially inform routing decisions. This contrasts with the recent findings of \citet{xue2024openmoe}, claiming that routing in decoder layers mainly depends on token identities and positions.

\begin{figure}[!tb]
    \includegraphics[width=0.5\textwidth]{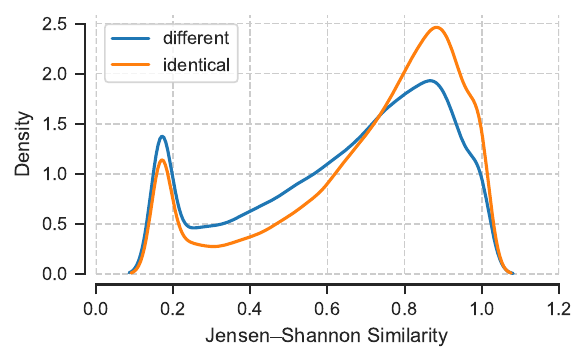}
    \caption{ \small Density estimates for routing similarities of ambiguous words given different and identical contexts. Routing decisions are aggregated across expert configurations. 
    }
    \label{fig:density}
\end{figure}

Figure \ref{fig:density} illustrates the \textit{Kernel Density Estimates} (KDE) for the routing similarities, distinguishing between word pairs stemming from identical contexts and those from different contexts of \texttt{WiC}. Note that the density estimates are calculated across expert configurations in $\{8, 16, 32, 64, 128\}$. 

The density curves are shaped similarly with a bimodal distribution, with density peaks at low and high values for the routing similarities visibly distinguishable. The density peak at high values indicates that, for many word pairs in identical contexts, the routing probabilities are quite similar, reflecting a commendable level of consistency in routing. The density peak at lower values suggests diverse routing patterns for many word pairs from different contexts, as desired when the context differs significantly. However, the overlap in the density curves implies that some word pairs receive similar routing despite having dissimilar meanings, which may occur in texts where the contexts are not substantially distinct, or the context differences are not clearly delineated by the language model.

\begin{figure}[!tb]
    \includegraphics[width=0.5\textwidth]{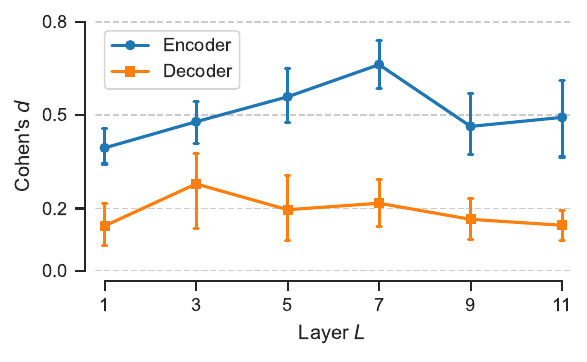}
    \caption{ \small
        Layer-wise effect sizes using Cohen's $d$ on the routing similarities of ambiguous words given some context. Routing decisions are aggregated across expert configurations. 
    }
    \label{fig:effects}
\end{figure}

Figure \ref{fig:effects} provides a layered investigation of the effect sizes of context sensitivity in the encoder and decoder layers. We measured the effect size using Cohen's $d$ by comparing the difference in routing similarities of words from identical and different contexts of \texttt{WiC}. We find that context is consistently significant for the routers in the encoder layers, whereas the routers in the decoder layers maintain a relatively stable and considerably lower effect sizes to context. Specifically, context integrates progressively in the early layers, peaks in the middle layers, and then slightly diminishes in the rear layers. This pattern can be attributed to late routers being specialized for span reconstruction.

\subsection{Correlation with Ambiguity}

\begin{figure}[!tb]
    \centering
    \includegraphics[width=0.5\textwidth]{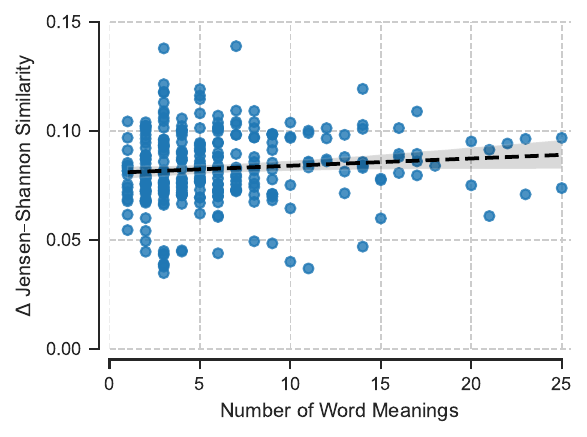}
    \caption{\small
        Differences in routing similarities for a set of ambiguous words given some context, as a function of the number of unique meanings derived from \texttt{WordNet}.
    }
    \label{fig:ambiguity}
\end{figure}

Since words can have multiple, potentially unrelated, meanings depending on the context, we are interested if routing decisions for ambiguous words vary with the number of meanings. Figure \ref{fig:ambiguity} plots differences in routing similarities against the number of word meanings derived from \texttt{WordNet} \citep{miller1995wordnet} \footnote{\texttt{WordNet} provides sets of synonyms that share a common meaning. To measure the number of meanings of a word, we counted the occurrence of a word in distinct \textit{synsets}.}. Although the trend line indicates that the context sensitivity of words correlates (insignificantly) with the number of distinct meanings, there is considerable variability, particularly for words with few meanings. This variability suggests that factors besides the number of meanings, such as word frequency, may determine the consistency of token-expert assignments in learned routers.


\section{Conclusion}

Given the claims surrounding the factors influencing routing decisions in sparsely-gated mixture-of-experts language models \citep{zoph2022st, xue2024openmoe}, we provide valuable insights into the influence of similarity and context. While similarity, encapsulated by token identities, form a stable basis for routing decisions, contextual cues provide an additional layer of refinement. However, the varying impact of context on the encoder and decoder reveals different sensitivities within the model components. The encoder demonstrates a strong ability to assign words in similar contexts consistently, revealing a high sensitivity to contextual cues, especially for configurations with many experts per sparse layer. The response of the decoder to context is poorer and more variable. This variability indicates instabilities in the utilization of context with respect to the number of experts.

Since our study demonstrates that context plays a significant role in routing, we hope that our approach sparks research on other linguistic properties and their influence on routing decisions, \textit{e.g.}, the influence of (affixal) negation \citep{vanson2016building} or the consistency of routing for multi-word expressions \citep{kochmar2020detecting}.

\paragraph{Limitation.} Challenging current claims about the context sensitivity of sparsely-gated language models, this study is limited by its focus on the \texttt{Switch} transformer model with its encoder-decoder architecture. Therefore, our findings may not be directly applicable to other types of transformer architectures, such as purely autoregressive models optimized with next-word prediction. We thus advocate for endeavors that expand the scope of analysis to cover a broader range of transformer architectures and develop more refined routing mechanisms to better integrate contextual cues, particularly for words with high polysemy.



\bibliography{acl_latex.bib}

\begin{thebibliography}{42}
\providecommand{\natexlab}[1]{#1}

\bibitem[{Bengio et~al.(2013)Bengio, L{\'e}onard, and Courville}]{bengio2013estimating}
Yoshua Bengio, Nicholas L{\'e}onard, and Aaron Courville. 2013.
\newblock Estimating or propagating gradients through stochastic neurons for conditional computation.
\newblock \emph{arXiv preprint arXiv:1308.3432}.

\bibitem[{Brown et~al.(2020)Brown, Mann, Ryder, Subbiah, Kaplan, Dhariwal, Neelakantan, Shyam, Sastry, Askell et~al.}]{brown2020language}
Tom Brown, Benjamin Mann, Nick Ryder, Melanie Subbiah, Jared~D Kaplan, Prafulla Dhariwal, Arvind Neelakantan, Pranav Shyam, Girish Sastry, Amanda Askell, et~al. 2020.
\newblock Language models are few-shot learners.
\newblock \emph{Advances in neural information processing systems}, 33:1877--1901.

\bibitem[{Chi et~al.(2022)Chi, Dong, Huang, Dai, Ma, Patra, Singhal, Bajaj, Song, Mao et~al.}]{chi2022representation}
Zewen Chi, Li~Dong, Shaohan Huang, Damai Dai, Shuming Ma, Barun Patra, Saksham Singhal, Payal Bajaj, Xia Song, Xian-Ling Mao, et~al. 2022.
\newblock On the representation collapse of sparse mixture of experts.
\newblock \emph{Advances in Neural Information Processing Systems}, 35:34600--34613.

\bibitem[{Clark et~al.(2022)Clark, de~Las~Casas, Guy, Mensch, Paganini, Hoffmann, Damoc, Hechtman, Cai, Borgeaud et~al.}]{clark2022unified}
Aidan Clark, Diego de~Las~Casas, Aurelia Guy, Arthur Mensch, Michela Paganini, Jordan Hoffmann, Bogdan Damoc, Blake Hechtman, Trevor Cai, Sebastian Borgeaud, et~al. 2022.
\newblock Unified scaling laws for routed language models.
\newblock In \emph{International conference on machine learning}, pages 4057--4086. PMLR.

\bibitem[{Dai et~al.(2022)Dai, Dong, Ma, Zheng, Sui, Chang, and Wei}]{dai2022stablemoe}
Damai Dai, Li~Dong, Shuming Ma, Bo~Zheng, Zhifang Sui, Baobao Chang, and Furu Wei. 2022.
\newblock \href {https://doi.org/10.18653/v1/2022.acl-long.489} {{S}table{M}o{E}: Stable routing strategy for mixture of experts}.
\newblock In \emph{Proceedings of the 60th Annual Meeting of the Association for Computational Linguistics (Volume 1: Long Papers)}, pages 7085--7095, Dublin, Ireland. Association for Computational Linguistics.

\bibitem[{Devlin et~al.(2019)Devlin, Chang, Lee, and Toutanova}]{devlin2019bert}
Jacob Devlin, Ming-Wei Chang, Kenton Lee, and Kristina Toutanova. 2019.
\newblock \href {https://doi.org/10.18653/v1/N19-1423} {{BERT}: Pre-training of deep bidirectional transformers for language understanding}.
\newblock In \emph{Proceedings of the 2019 Conference of the North {A}merican Chapter of the Association for Computational Linguistics: Human Language Technologies, Volume 1 (Long and Short Papers)}, pages 4171--4186, Minneapolis, Minnesota. Association for Computational Linguistics.

\bibitem[{Do et~al.(2023)Do, Le, Pham, Nguyen, Doan, Nguyen, Liu, Ramasamy, Li, and Hoi}]{do2023hyperrouter}
Giang Do, Khiem Le, Quang Pham, Trungtin Nguyen, Thanh-Nam Doan, Bint~T Nguyen, Chenghao Liu, Savitha Ramasamy, Xiaoli Li, and Steven Hoi. 2023.
\newblock Hyperrouter: Towards efficient training and inference of sparse mixture of experts.
\newblock \emph{arXiv preprint arXiv:2312.07035}.

\bibitem[{Du et~al.(2022)Du, Huang, Dai, Tong, Lepikhin, Xu, Krikun, Zhou, Yu, Firat et~al.}]{du2022glam}
Nan Du, Yanping Huang, Andrew~M Dai, Simon Tong, Dmitry Lepikhin, Yuanzhong Xu, Maxim Krikun, Yanqi Zhou, Adams~Wei Yu, Orhan Firat, et~al. 2022.
\newblock Glam: Efficient scaling of language models with mixture-of-experts.
\newblock In \emph{International Conference on Machine Learning}, pages 5547--5569. PMLR.

\bibitem[{Fedus et~al.(2022)Fedus, Zoph, and Shazeer}]{fedus2022switch}
William Fedus, Barret Zoph, and Noam Shazeer. 2022.
\newblock Switch transformers: Scaling to trillion parameter models with simple and efficient sparsity.
\newblock \emph{Journal of Machine Learning Research}, 23(120):1--39.

\bibitem[{Finkelstein et~al.(2001)Finkelstein, Gabrilovich, Matias, Rivlin, Solan, Wolfman, and Ruppin}]{finkelstein2001placing}
Lev Finkelstein, Evgeniy Gabrilovich, Yossi Matias, Ehud Rivlin, Zach Solan, Gadi Wolfman, and Eytan Ruppin. 2001.
\newblock Placing search in context: The concept revisited.
\newblock In \emph{Proceedings of the 10th international conference on World Wide Web}, pages 406--414.

\bibitem[{Gale et~al.(2023)Gale, Narayanan, Young, and Zaharia}]{gale2023megablocks}
Trevor Gale, Deepak Narayanan, Cliff Young, and Matei Zaharia. 2023.
\newblock Megablocks: Efficient sparse training with mixture-of-experts.
\newblock \emph{Proceedings of Machine Learning and Systems}, 5:288--304.

\bibitem[{Hill et~al.(2015)Hill, Reichart, and Korhonen}]{hill2015simlex}
Felix Hill, Roi Reichart, and Anna Korhonen. 2015.
\newblock Simlex-999: Evaluating semantic models with (genuine) similarity estimation.
\newblock \emph{Computational Linguistics}, 41(4):665--695.

\bibitem[{Huang et~al.(2012)Huang, Socher, Manning, and Ng}]{huang2012improving}
Eric~H Huang, Richard Socher, Christopher~D Manning, and Andrew~Y Ng. 2012.
\newblock Improving word representations via global context and multiple word prototypes.
\newblock In \emph{Proceedings of the 50th annual meeting of the association for computational linguistics (Volume 1: Long papers)}, pages 873--882.

\bibitem[{Jacobs et~al.(1991)Jacobs, Jordan, Nowlan, and Hinton}]{jacobs1991adaptive}
Robert~A Jacobs, Michael~I Jordan, Steven~J Nowlan, and Geoffrey~E Hinton. 1991.
\newblock Adaptive mixtures of local experts.
\newblock \emph{Neural computation}, 3(1):79--87.

\bibitem[{Jiang et~al.(2024)Jiang, Sablayrolles, Roux, Mensch, Savary, Bamford, Chaplot, Casas, Hanna, Bressand et~al.}]{jiang2024mixtral}
Albert~Q Jiang, Alexandre Sablayrolles, Antoine Roux, Arthur Mensch, Blanche Savary, Chris Bamford, Devendra~Singh Chaplot, Diego de~las Casas, Emma~Bou Hanna, Florian Bressand, et~al. 2024.
\newblock Mixtral of experts.
\newblock \emph{arXiv preprint arXiv:2401.04088}.

\bibitem[{Kaplan et~al.(2020)Kaplan, McCandlish, Henighan, Brown, Chess, Child, Gray, Radford, Wu, and Amodei}]{kaplan2020scaling}
Jared Kaplan, Sam McCandlish, Tom Henighan, Tom~B Brown, Benjamin Chess, Rewon Child, Scott Gray, Alec Radford, Jeffrey Wu, and Dario Amodei. 2020.
\newblock Scaling laws for neural language models.
\newblock \emph{arXiv preprint arXiv:2001.08361}.

\bibitem[{Kochmar et~al.(2020)Kochmar, Gooding, and Shardlow}]{kochmar2020detecting}
Ekaterina Kochmar, Sian Gooding, and Matthew Shardlow. 2020.
\newblock \href {https://aclanthology.org/2020.lrec-1.545} {Detecting multiword expression type helps lexical complexity assessment}.
\newblock In \emph{Proceedings of the Twelfth Language Resources and Evaluation Conference}, pages 4426--4435, Marseille, France. European Language Resources Association.

\bibitem[{Komatsuzaki et~al.(2022)Komatsuzaki, Puigcerver, Lee-Thorp, Ruiz, Mustafa, Ainslie, Tay, Dehghani, and Houlsby}]{komatsuzaki2022sparse}
Aran Komatsuzaki, Joan Puigcerver, James Lee-Thorp, Carlos~Riquelme Ruiz, Basil Mustafa, Joshua Ainslie, Yi~Tay, Mostafa Dehghani, and Neil Houlsby. 2022.
\newblock Sparse upcycling: Training mixture-of-experts from dense checkpoints.
\newblock \emph{arXiv preprint arXiv:2212.05055}.

\bibitem[{Kudo and Richardson(2018)}]{kudo2018sentencepiece}
Taku Kudo and John Richardson. 2018.
\newblock \href {https://doi.org/10.18653/v1/D18-2012} {{S}entence{P}iece: A simple and language independent subword tokenizer and detokenizer for neural text processing}.
\newblock In \emph{Proceedings of the 2018 Conference on Empirical Methods in Natural Language Processing: System Demonstrations}, pages 66--71, Brussels, Belgium. Association for Computational Linguistics.

\bibitem[{Lepikhin et~al.(2020)Lepikhin, Lee, Xu, Chen, Firat, Huang, Krikun, Shazeer, and Gshard}]{lepikhin2020scaling}
D~Lepikhin, H~Lee, Y~Xu, D~Chen, O~Firat, Y~Huang, M~Krikun, N~Shazeer, and Z~Gshard. 2020.
\newblock Scaling giant models with conditional computation and automatic sharding.
\newblock \emph{arXiv preprint arXiv:2006.16668}.

\bibitem[{Lewis et~al.(2021)Lewis, Bhosale, Dettmers, Goyal, and Zettlemoyer}]{lewis2021base}
Mike Lewis, Shruti Bhosale, Tim Dettmers, Naman Goyal, and Luke Zettlemoyer. 2021.
\newblock Base layers: Simplifying training of large, sparse models.
\newblock In \emph{International Conference on Machine Learning}, pages 6265--6274. PMLR.

\bibitem[{Li et~al.(2023)Li, Su, Yang, Jiang, Wang, and Xu}]{li2023adaptive}
Jiamin Li, Qiang Su, Yitao Yang, Yimin Jiang, Cong Wang, and Hong Xu. 2023.
\newblock \href {https://doi.org/10.18653/v1/2023.emnlp-main.217} {Adaptive gating in mixture-of-experts based language models}.
\newblock In \emph{Proceedings of the 2023 Conference on Empirical Methods in Natural Language Processing}, pages 3577--3587, Singapore. Association for Computational Linguistics.

\bibitem[{Lin et~al.(2024)Lin, Tang, Ye, Cui, Zhu, Jin, Zhang, Ning, and Yuan}]{lin2024moe}
Bin Lin, Zhenyu Tang, Yang Ye, Jiaxi Cui, Bin Zhu, Peng Jin, Junwu Zhang, Munan Ning, and Li~Yuan. 2024.
\newblock Moe-llava: Mixture of experts for large vision-language models.
\newblock \emph{arXiv preprint arXiv:2401.15947}.

\bibitem[{Liu et~al.(2022)Liu, Kim, Muzio, and Hassan}]{liu2022gating}
Rui Liu, Young~Jin Kim, Alexandre Muzio, and Hany Hassan. 2022.
\newblock Gating dropout: Communication-efficient regularization for sparsely activated transformers.
\newblock In \emph{International Conference on Machine Learning}, pages 13782--13792. PMLR.

\bibitem[{Miller(1995)}]{miller1995wordnet}
George~A Miller. 1995.
\newblock Wordnet: a lexical database for english.
\newblock \emph{Communications of the ACM}, 38(11):39--41.

\bibitem[{Nie et~al.(2021)Nie, Miao, Cao, Ma, Liu, Xue, Miao, Liu, Yang, and Cui}]{nie2021evomoe}
Xiaonan Nie, Xupeng Miao, Shijie Cao, Lingxiao Ma, Qibin Liu, Jilong Xue, Youshan Miao, Yi~Liu, Zhi Yang, and Bin Cui. 2021.
\newblock Evomoe: An evolutional mixture-of-experts training framework via dense-to-sparse gate.
\newblock \emph{arXiv preprint arXiv:2112.14397}.

\bibitem[{Pilehvar and Camacho-Collados(2019)}]{pilehvar2019wic}
Mohammad~Taher Pilehvar and Jose Camacho-Collados. 2019.
\newblock \href {https://doi.org/10.18653/v1/N19-1128} {{W}i{C}: the word-in-context dataset for evaluating context-sensitive meaning representations}.
\newblock In \emph{Proceedings of the 2019 Conference of the North {A}merican Chapter of the Association for Computational Linguistics: Human Language Technologies, Volume 1 (Long and Short Papers)}, pages 1267--1273, Minneapolis, Minnesota. Association for Computational Linguistics.

\bibitem[{Puigcerver et~al.(2024)Puigcerver, Ruiz, Mustafa, and Houlsby}]{puigcerver2024sparse}
Joan Puigcerver, Carlos~Riquelme Ruiz, Basil Mustafa, and Neil Houlsby. 2024.
\newblock From sparse to soft mixtures of experts.
\newblock In \emph{The Twelfth International Conference on Learning Representations}.

\bibitem[{Radford et~al.(2019)Radford, Wu, Child, Luan, Amodei, Sutskever et~al.}]{radford2019language}
Alec Radford, Jeffrey Wu, Rewon Child, David Luan, Dario Amodei, Ilya Sutskever, et~al. 2019.
\newblock Language models are unsupervised multitask learners.
\newblock \emph{OpenAI blog}, 1(8):9.

\bibitem[{Raffel et~al.(2020)Raffel, Shazeer, Roberts, Lee, Narang, Matena, Zhou, Li, and Liu}]{raffel2020exploring}
Colin Raffel, Noam Shazeer, Adam Roberts, Katherine Lee, Sharan Narang, Michael Matena, Yanqi Zhou, Wei Li, and Peter~J Liu. 2020.
\newblock Exploring the limits of transfer learning with a unified text-to-text transformer.
\newblock \emph{Journal of machine learning research}, 21(140):1--67.

\bibitem[{Riquelme et~al.(2021)Riquelme, Puigcerver, Mustafa, Neumann, Jenatton, Susano~Pinto, Keysers, and Houlsby}]{riquelme2021scaling}
Carlos Riquelme, Joan Puigcerver, Basil Mustafa, Maxim Neumann, Rodolphe Jenatton, Andr{\'e} Susano~Pinto, Daniel Keysers, and Neil Houlsby. 2021.
\newblock Scaling vision with sparse mixture of experts.
\newblock \emph{Advances in Neural Information Processing Systems}, 34:8583--8595.

\bibitem[{Roller et~al.(2021)Roller, Sukhbaatar, Weston et~al.}]{roller2021hash}
Stephen Roller, Sainbayar Sukhbaatar, Jason Weston, et~al. 2021.
\newblock Hash layers for large sparse models.
\newblock \emph{Advances in Neural Information Processing Systems}, 34:17555--17566.

\bibitem[{Shazeer et~al.(2017)Shazeer, Mirhoseini, Maziarz, Davis, Le, Hinton, and Dean}]{shazeer2017outrageously}
Noam Shazeer, Azalia Mirhoseini, Krzysztof Maziarz, Andy Davis, Quoc Le, Geoffrey Hinton, and Jeff Dean. 2017.
\newblock Outrageously large neural networks: The sparsely-gated mixture-of-experts layer.
\newblock \emph{arXiv preprint arXiv:1701.06538}.

\bibitem[{Shen et~al.(2023)Shen, Yao, Li, Darrell, Keutzer, and He}]{shen2023scaling}
Sheng Shen, Zhewei Yao, Chunyuan Li, Trevor Darrell, Kurt Keutzer, and Yuxiong He. 2023.
\newblock \href {https://doi.org/10.18653/v1/2023.findings-emnlp.758} {Scaling vision-language models with sparse mixture of experts}.
\newblock In \emph{Findings of the Association for Computational Linguistics: EMNLP 2023}, pages 11329--11344, Singapore. Association for Computational Linguistics.

\bibitem[{Touvron et~al.(2023)Touvron, Lavril, Izacard, Martinet, Lachaux, Lacroix, Rozi{\`e}re, Goyal, Hambro, Azhar et~al.}]{touvron2023llama}
Hugo Touvron, Thibaut Lavril, Gautier Izacard, Xavier Martinet, Marie-Anne Lachaux, Timoth{\'e}e Lacroix, Baptiste Rozi{\`e}re, Naman Goyal, Eric Hambro, Faisal Azhar, et~al. 2023.
\newblock Llama: Open and efficient foundation language models.
\newblock \emph{arXiv preprint arXiv:2302.13971}.

\bibitem[{van Son et~al.(2016)van Son, van Miltenburg, and Morante}]{vanson2016building}
Chantal van Son, Emiel van Miltenburg, and Roser Morante. 2016.
\newblock \href {https://aclanthology.org/W16-5007} {Building a dictionary of affixal negations}.
\newblock In \emph{Proceedings of the Workshop on Extra-Propositional Aspects of Meaning in Computational Linguistics ({E}x{P}ro{M})}, pages 49--56, Osaka, Japan. The COLING 2016 Organizing Committee.

\bibitem[{Vaswani et~al.(2017)Vaswani, Shazeer, Parmar, Uszkoreit, Jones, Gomez, Kaiser, and Polosukhin}]{vaswani2017attention}
Ashish Vaswani, Noam Shazeer, Niki Parmar, Jakob Uszkoreit, Llion Jones, Aidan~N Gomez, {\L}ukasz Kaiser, and Illia Polosukhin. 2017.
\newblock Attention is all you need.
\newblock \emph{Advances in neural information processing systems}, 30.

\bibitem[{Xue et~al.(2024)Xue, Zheng, Fu, Ni, Zheng, Zhou, and You}]{xue2024openmoe}
Fuzhao Xue, Zian Zheng, Yao Fu, Jinjie Ni, Zangwei Zheng, Wangchunshu Zhou, and Yang You. 2024.
\newblock Openmoe: An early effort on open mixture-of-experts language models.
\newblock \emph{arXiv preprint arXiv:2402.01739}.

\bibitem[{Yang et~al.(2021)Yang, Lin, Men, Zhou, Jiang, Jia, Wang, Zhang, Wang, Li et~al.}]{yang2021m6}
An~Yang, Junyang Lin, Rui Men, Chang Zhou, Le~Jiang, Xianyan Jia, Ang Wang, Jie Zhang, Jiamang Wang, Yong Li, et~al. 2021.
\newblock M6-t: Exploring sparse expert models and beyond.
\newblock \emph{arXiv preprint arXiv:2105.15082}.

\bibitem[{Zhou et~al.(2022)Zhou, Lei, Liu, Du, Huang, Zhao, Dai, Le, Laudon et~al.}]{zhou2022mixture}
Yanqi Zhou, Tao Lei, Hanxiao Liu, Nan Du, Yanping Huang, Vincent Zhao, Andrew~M Dai, Quoc~V Le, James Laudon, et~al. 2022.
\newblock Mixture-of-experts with expert choice routing.
\newblock \emph{Advances in Neural Information Processing Systems}, 35:7103--7114.

\bibitem[{Zoph et~al.(2022)Zoph, Bello, Kumar, Du, Huang, Dean, Shazeer, and Fedus}]{zoph2022st}
Barret Zoph, Irwan Bello, Sameer Kumar, Nan Du, Yanping Huang, Jeff Dean, Noam Shazeer, and William Fedus. 2022.
\newblock St-moe: Designing stable and transferable sparse expert models.
\newblock \emph{arXiv preprint arXiv:2202.08906}.

\bibitem[{Zuo et~al.(2021)Zuo, Liu, Jiao, Kim, Hassan, Zhang, Zhao, and Gao}]{zuo2021taming}
Simiao Zuo, Xiaodong Liu, Jian Jiao, Young~Jin Kim, Hany Hassan, Ruofei Zhang, Tuo Zhao, and Jianfeng Gao. 2021.
\newblock Taming sparsely activated transformer with stochastic experts.
\newblock \emph{arXiv preprint arXiv:2110.04260}.

\end{thebibliography}

\comment{
\appendix
\section{Appendix}
\label{sec:appendix}


\begin{figure*}[h]


    \subfigure[...]
    {
        \includegraphics[width=\textwidth]{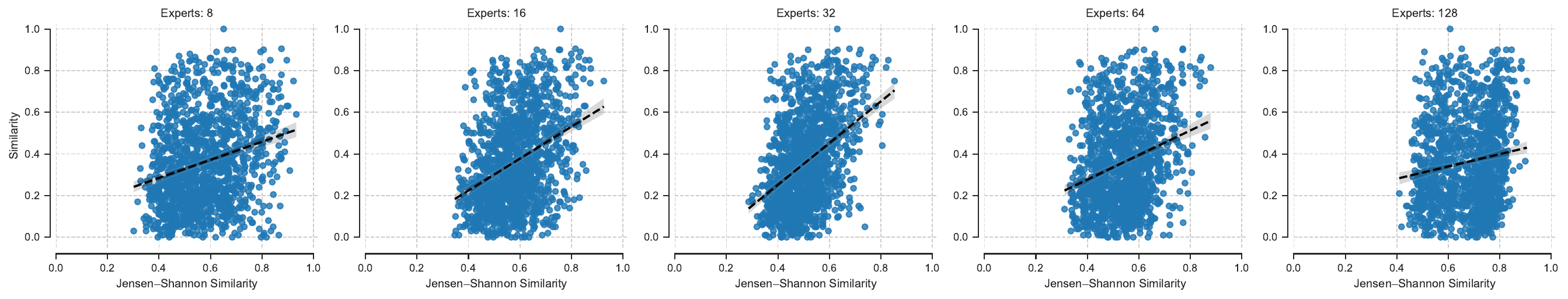}
        \label{fig:a}
    }
    \hfill
    \subfigure[...]
    {
        \includegraphics[width=\textwidth]{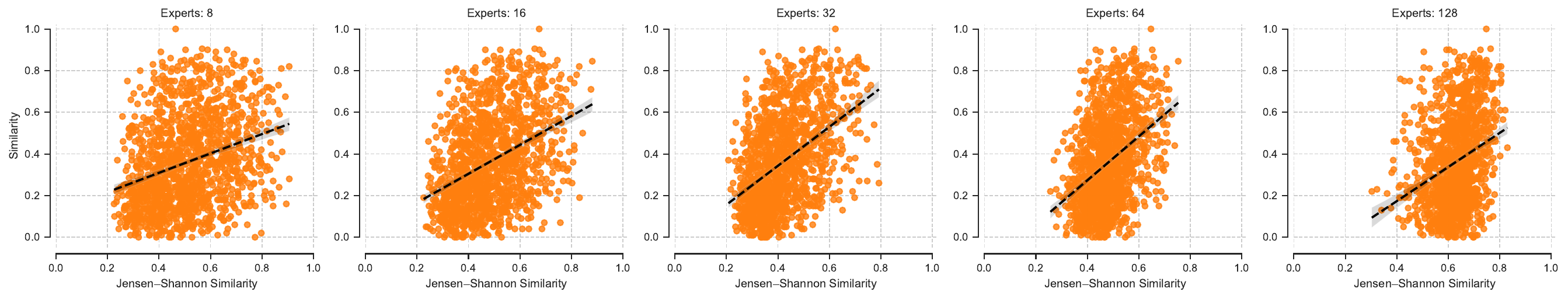}
        \label{fig:b}
    }
    \subfigure[...]
    {
        \includegraphics[width=\textwidth]{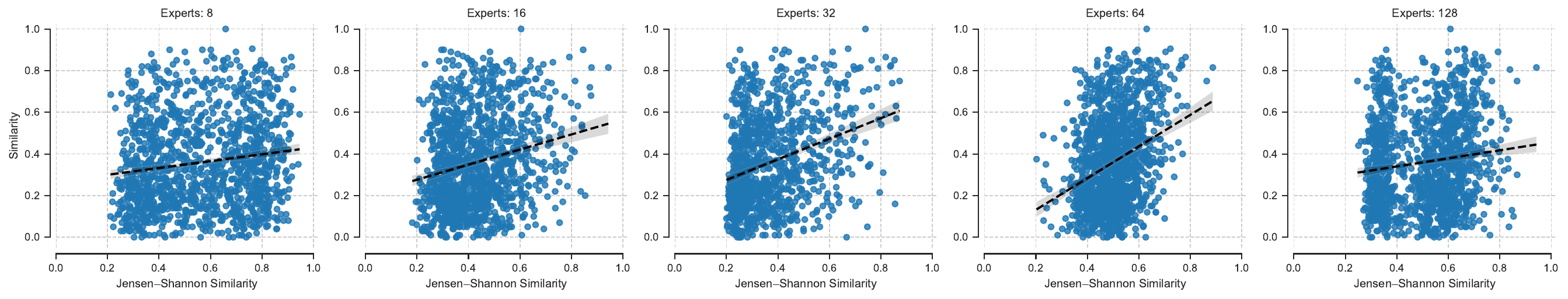}
        \label{fig:c}
    }
    \hfill
    \subfigure[...]
    {
        \includegraphics[width=\textwidth]{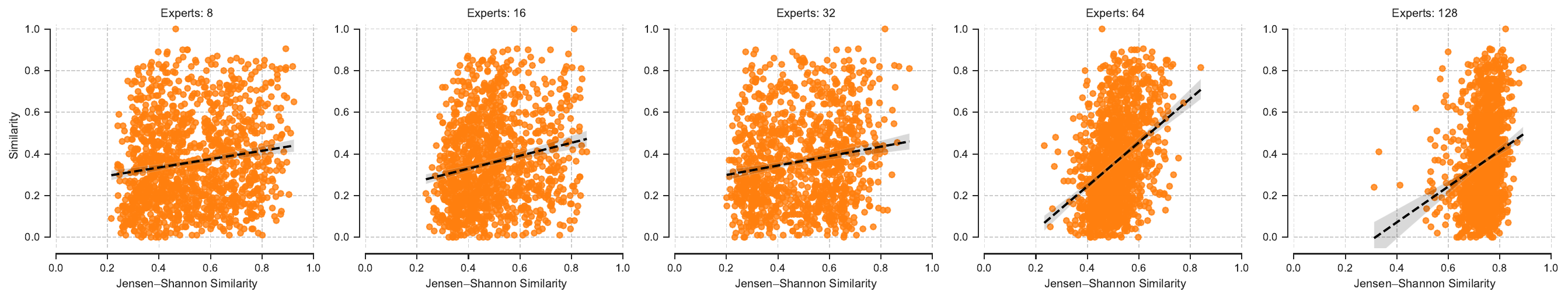}
        \label{fig:d}
    }
    \caption{
        ...
    }
    \label{fig:correlations}
\end{figure*}
}

\end{document}